\UseRawInputEncoding
\documentclass[lettersize,journal]{IEEEtran}
\usepackage{amsmath,amsfonts}
\usepackage{algorithmic}
\usepackage{array}
\usepackage{textcomp}
\usepackage{stfloats}
\usepackage{url}
\usepackage{verbatim}
\usepackage{graphicx}
\def\BibTeX{{\rm B\kern-.05em{\sc i\kern-.025em b}\kern-.08em
    T\kern-.1667em\lower.7ex\hbox{E}\kern-.125emX}}
\usepackage{balance}
\usepackage{enumitem}

\usepackage{amssymb}
\usepackage{booktabs}
\usepackage{bbm}
\usepackage{multirow}
\usepackage{siunitx}
\usepackage{caption}
\usepackage{subcaption}

\newenvironment{customenum}{
  \begin{enumerate}[label={}, leftmargin=*, align=left, itemsep=3pt, parsep=3pt]
}{
  \end{enumerate}
}

\begin{document}

\title{Probabilistic Machine Learning for Noisy Labels in Earth Observation}

\author{Spyros Kondylatos, Nikolaos Ioannis Bountos, Ioannis Prapas, Angelos Zavras, Gustau Camps-Valls \IEEEmembership{Fellow, IEEE}, Ioannis Papoutsis\thanks{S. Kondylatos is with Orion Lab, National Observatory of Athens, National Technical University of Athens, and Image Processing Laboratory (IPL), Universitat de Val\`encia\\
N.I. Bountos is with Orion Lab, National Observatory of Athens, National Technical University of Athens and the Harokopio University of Athens \\
I. Prapas is with Orion Lab, National Observatory of Athens, National Technical University of Athens, and Image Processing Laboratory (IPL), Universitat de Val\`encia \\
A. Zavras is with Orion Lab, National Observatory of Athens, National Technical University of Athens, and Harokopio University of Athens \\
G. Camps-Valls is with Image Processing Laboratory (IPL), Universitat de Val\`encia\\
I. Papoutsis is with Orion Lab, National Observatory of Athens, National Technical University of Athens and the Archimedes, Athena Research Center}}

\markboth{Journal of \LaTeX\ Class Files,~Vol.~18, No.~9, September~2020}%
{How to Use the IEEEtran \LaTeX \ Templates}

\maketitle

\begin{abstract}

Label noise poses a significant challenge in Earth Observation (EO), often degrading the performance and reliability of supervised Machine Learning (ML) models. 
Yet, given the critical nature of several EO applications, developing robust and trustworthy ML solutions is essential.
In this study, we take a step in this direction by leveraging probabilistic ML to model input-dependent label noise and quantify data uncertainty in EO tasks, accounting for the unique noise sources inherent in the domain.
We train uncertainty-aware probabilistic models across a broad range of high-impact EO applications---spanning diverse noise sources, input modalities, and ML configurations---and introduce a dedicated pipeline to assess their accuracy and reliability.
Our experimental results show that the uncertainty-aware models consistently outperform the standard deterministic approaches across most datasets and evaluation metrics.
Moreover, through rigorous uncertainty evaluation, we validate the reliability of the predicted uncertainty estimates, enhancing the interpretability of model predictions.  
Our findings emphasize the importance of modeling label noise and incorporating uncertainty quantification in EO, paving the way for more accurate, reliable, and trustworthy ML solutions in the field.

\end{abstract}


\section{Introduction}
\label{sec:intro}

\IEEEPARstart{T}{he} increasing availability of Remote Sensing (RS) data has significantly advanced research and applications across various scientific fields \cite{soille_versatile_2018}. 
Satellite-based Earth Observation (EO) has transformed the analysis and modeling of the Earth system, enabling continuous monitoring of environmental, natural, and human-driven processes of the globe \cite{reichstein_deep_2019}.
By providing large-scale data from space, EO offers critical insights into the dynamics and interactions of the Earth's five major spheres---Atmosphere, Hydrosphere, Geosphere, Biosphere, and Cryosphere.
This wealth of data supports a wide array of applications, such as climate change \cite{yang2013role}, food insecurity \cite{nakalembe2020urgent}, ecosystem monitoring \cite{wu_deep_2023}, and Earth Science applications \cite{dl_earth_sciences_camps-valls}, fostering a deeper understanding of the complex systems that shape our planet.

Deep Learning (DL) has emerged as a powerful tool for extracting knowledge from EO data.
DL methods have been successfully applied across a wide range of EO modalities, including optical imagery, Synthetic Aperture Radar (SAR), Interferometric Synthetic Aperture Radar (InSAR), and multi-modal data \cite{zhu_deep_2017}. 
Supervised DL models, trained on annotated datasets from these sources, have demonstrated remarkable performance in addressing several EO and geoscientific problems \cite{ma_deep_2019, Camps-Valls:291513}, showcasing great results across different applications \cite{zhang_deep_2016, aleissaee_transformers_2022}. 
However, despite their widespread adoption within the EO community, the effectiveness of supervised DL approaches remains highly dependent on the availability of large, high-quality labeled datasets \cite{lecun_deep_2015}. 

Annotating EO data presents unique challenges, often requiring expert domain knowledge tailored to specific tasks and applications; yet, manual annotation of large-scale datasets by domain experts is both costly and time-consuming.
To overcome these limitations, alternative strategies such as crowdsourcing, semi-supervised learning, or annotations by ML practitioners are frequently used \cite{sumbul_generative_2023}. 
While these methods improve scalability and reduce costs, they also introduce a higher risk of labeling errors, potentially compromising the quality of supervised datasets.
Even when expert annotations are employed, the process remains complex due to the intricate nature of EO applications.
These applications typically require a deep understanding of dynamic Earth processes, as well as the ability to disentangle complex signals and interpret natural phenomena from satellite imagery.  
As a result, the label generation process becomes uncertain \cite{elmes_accounting_2020}, often resulting in label noise within EO supervised datasets.
In single-label classification, label noise occurs when a sample is given the wrong label. 
In multi-label classification, it can appear as either missing labels, where a relevant label is not assigned to a sample, or incorrect labels, where a sample is mistakenly linked to a label that does not actually apply.
Label noise can also result from a combination of both missing and incorrect labels \cite{sumbul_generative_2023}.
Addressing label noise is crucial, as its presence can degrade the performance of supervised models, ultimately leading to less reliable predictions \cite{jiang_mentornet_2018}.


Reliability is a fundamental requirement for EO DL models \cite{gawlikowski_survey_2021, tuia_toward_2021}.
This is particularly relevant for EO applications related to disaster management, such as wildfire forecasting \cite{kondylatos_wildfire_2022}, flood mapping \cite{bountos2025kuro}, landslide monitoring \cite{boehm_deep_2022}, and volcanic unrest detection \cite{bountos_hephaestus_2022}.
These applications demand swift and precise decision-making, as poor choices can lead to loss of lives or environmental damage.
A key approach for improving the reliability of DL models relies on quantifying the uncertainty in their predictions. 
Assessing the confidence level of model outputs can enhance decision-making, allowing stakeholders to adapt their strategies accordingly.
For instance, in volcanic activity early-warning systems, emergency response protocols may vary depending on the model's confidence.
A highly certain prediction of an imminent eruption may initiate large-scale evacuations and resource allocation. 
Conversely, a high-uncertainty prediction may lead decision-makers to opt for closer monitoring and precautionary measures rather than immediate, large-scale interventions.

In Machine Learning (ML), uncertainty is typically classified into two main types: epistemic (model) uncertainty and aleatoric (data) uncertainty. 
Epistemic uncertainty arises from a model's lack of knowledge and can be reduced as more data becomes available or the model improves. 
On the other hand, aleatoric uncertainty stems from inherent noise in the data and cannot be reduced, even with additional data samples.
Aleatoric uncertainty can be further categorized into homoscedastic (input-independent), where the noise remains constant across all samples in the input space, and heteroscedastic (input-dependent), where the level of noise varies depending on the input \cite{Gal2016UncertaintyID, kendall_what_2017}.

The relationship between label noise and heteroscedastic aleatoric uncertainty is significant as the latter increases in samples that are more challenging to annotate.
This connection has been explored within the Computer Vision (CV) community, leading to the development of probabilistic models that estimate heteroscedastic aleatoric uncertainty by modeling input-dependent label noise \cite{kendall_what_2017, collier_simple_2020}.
In these models, a distribution is typically placed over the logits of a Neural Network (NN), where the mean represents the predicted output and the variance quantifies the associated uncertainty.
These methods have improved model performance and enabled the quantification of data uncertainty across various CV tasks impacted by label noise.
To the best of our knowledge, such methods have yet to be explored in the EO domain.

The widespread presence of label noise, coupled with the growing need for reliable uncertainty estimation in EO tasks, highlights the potential of such methods in EO.
In this work, we adopt the probabilistic framework introduced by Collier et al. \cite{collier_simple_2020} to model input-dependent label noise and quantify data uncertainty in EO tasks.
To systematically assess the framework's effectiveness in handling the diverse sources of label noise specific to the domain, we apply it to four EO applications, each characterized by a distinct source of label noise, varying input modalities, and ML setups. 
We assess its predictive performance against standard deterministic DL models and leverage its probabilistic nature to quantify estimates of the aleatoric uncertainty.
To examine the practical utility of these uncertainties, we use two complementary approaches: i) we assess their reliability through dedicated evaluation strategies, and ii) we use the predicted uncertainty estimates to identify certain and uncertain samples across the various EO tasks, demonstrating their actual applicability in real-world applications.

The main contributions of this work can be summarized as follows:
\begin{itemize}
\item We provide a comprehensive classification of label noise sources specific to EO datasets.
\item We investigate the application of probabilistic ML to mitigate the impact of label noise in EO, demonstrating its effectiveness across four high-stakes applications.
\item We propose a dedicated pipeline for integrating uncertainty-aware ML in EO, incorporating uncertainty quantification, evaluation, and visualization. This framework highlights the critical role of uncertainty in enhancing model reliability and supporting informed decision-making in critical EO tasks.
\end{itemize}

\section{Background \& Related Work} 

\subsection{Learning under Label Noise in Deep Learning}

\noindent In supervised DL, a finite set of training data denoted as: $$D = \{(x_1, y_1), (x_2, y_2), \ldots (x_n, y_n)\} = X \times Y$$ is used for training, with $X$ representing the input data and $Y$ the corresponding labels.
A noisy training dataset is typically defined as:
$$D = X \times \tilde{Y},$$ where $\tilde{Y}$ represents the set of observed labels that may differ from the true labels $Y$.
Various strategies have been proposed to mitigate the effects of label noise in DL \cite{cordeiro_survey_2020}, which can be broadly categorized into four main approaches: \textit{Robust Architecture, Robust Regularization, Robust Losses, and Sample Selection} \cite{song_learning_2022}.

\textit{Robust Architectures} modify NN structure to account for noisy labels.
One common approach involves the integration of noise adaptation layers, which estimate a label transition matrix $T$ that models the probability of a clean label being flipped into a noisy one, defined as $T_{ij} = p(\tilde{y}=j | y=i,x)$ \cite{patrini_making_2017, hendrycks_using_2019, goldberger_training_2022}. 
\textit{Robust Regularization} techniques focus on reducing overfitting to noisy labels.
Explicit regularization strategies, involve regularizing loss functions to discourage memorization of noise \cite{tanno_learning_2019, jenni_deep_2018, menon_2020, wei_open-set_2021}, while implicit regularization includes adversarial training \cite{goodfellow_explaining_2015}, label smoothing \cite{pereyra_regularizing_2017, lukasik_does_2020} and mixup \cite{zhang_mixup_2018}.
\textit{Robust Losses} introduce noise-tolerant formulations that provide greater resilience to corrupted labels.
Such losses include mean absolute error \cite{ghosh_robust_2017}, generalized cross-entropy \cite{zhang_generalized_2018}, and symmetric cross-entropy loss \cite{wang_symmetric_2019}.
Additionally, loss correction techniques \cite{hendrycks_using_2019, zhang_approximating_2021, yang_estimating_2022} and reweighting \cite{wang_multiclass_2018, chang_active_2017, zhang_dualgraph_2021} adjust the loss dynamically to mitigate the impact of label noise.
A related approach, label refurbishment, adjusts the loss by incorporating refurbished labels obtained from a convex combination of noisy and predicted labels \cite{ma_dimensionality-driven_2018, huang_self-adaptive_2020,chen_beyond_2020, song_selfie_2019}. 
Finally, \textit{Sample Selection} methods focus on identifying clean samples from noisy datasets, often leveraging co-teaching networks \cite{jiang_mentornet_2018, han_co-teaching_2018, yu_how_2019, wei_combating_2020} or iterative training strategies that dynamically filter out unreliable samples during training \cite{shen_learning_2019, chen_understanding_2019, huang_o2u-net_2019}.

While many approaches have been proposed to address label noise in DL, the critical nature of EO applications demands approaches that go beyond improving model performance to ensure practical reliability.
In this regard, uncertainty estimation plays a crucial role.
Therefore, this work focuses on methods that jointly address label noise and incorporate uncertainty estimation, ensuring both robust and trustworthy modeling.

\subsection{Uncertainty in Supervised Deep Learning}

\noindent Considering the set of training data: $$D = X \times Y,$$ a DL model $f^w$, parameterized by weights $w$ is typically trained to map $f^w(x) = y$ for each pair $(x, y)$ in $X \times Y$.
Given a new sample $x^*$, classical deterministic DL models provide a single-point estimate $f^w(x^*) = y^*$, without accounting for the inherent uncertainty both in the data and the model itself. 
However, predictive uncertainty in DL classification arises from multiple factors \cite{gawlikowski_survey_2021}, including: \textit{i) variability in real-world situations, ii) errors introduced by the measurement systems or labeling process, iii) suboptimal model architecture design, iv) inaccuracies in the training procedure of the DL model, and v) the presence of previously unseen data.}
Epistemic uncertainty is associated with factors i, iii, iv, and v, and aleatoric uncertainty arises from factor ii \cite{hullermeier_aleatoric_2021}.


\textit{Epistemic uncertainty} in DL is commonly captured through Bayesian Neural Networks (BNNs) \cite{jospin_hands-bayesian_2020} and Deep Ensembles (DEs) \cite{lakshminarayanan_simple_2017}. 
BNNs integrate a prior probability distribution into network parameters and leverage Bayesian techniques for posterior estimation, using methods like Markov Chain Monte Carlo  \cite{bishop_pattern_2009, welling_bayesian_nodate} or Variational Inference \cite{barber_ensemble_nodate, blundell_weight_2015, hernandez-lobato_probabilistic_2015, gal_dropout_2016}. 
DEs provide uncertainty estimates by aggregating predictions from multiple independently trained deterministic NNs.

\noindent\textit{Aleatoric uncertainty} is calculated using:
i) external methods, applied to an already trained deterministic NN \cite{ramalho_density_2019, oberdiek_classification_2018, lee_gradients_2020}, ii) network modifications, where the architecture is adapted to predict uncertainty directly during training, often by learning the parameters of a probabilistic distribution in the output space \cite{malinin_predictive_2018, sensoy_evidential_2018, nandy_towards_2021}, and iii) test-time data augmentation, where the model generates multiple predictions by augmenting inputs at inference time and calculate uncertainty as their variability  \cite{ayhan_test-time_2018, wang_aleatoric_2019}.

Kendall and Gal \cite{kendall_what_2017} introduced a network modification method for quantifying heteroscedastic aleatoric uncertainty by explicitly modeling label noise. 
In this method, a Gaussian distribution is placed on the logits of a standard softmax classification model before applying the softmax function.
Formally, given an input $x$, the logits $u_{c}(x)$ are modeled as:
\begin{equation*}
\begin{aligned}
  u_{c} (x) \sim \mathcal{N} (f_{c}^{w}(x), \sigma_{c}^{w}(x)^2), \forall c = 1,\ldots, K,
\end{aligned}
  \label{eq:tau=2}
\end{equation*}
where $f_{c}^{w}(x)$ represents the mean logit prediction and $\sigma_{c}^{w}(x)^2$ the variance, both of which are learned by the model.
The final class probabilities are then obtained via the softmax function:
\begin{equation}
\begin{aligned}
  p_c (x) = \frac{\exp{(u_c(x))}}{\sum_{k=1}^{K}\exp(u_k(x))}, \forall c = 1,\ldots, K.
\end{aligned}
  \label{eq:tau=1}
\end{equation}
where $K$ is the total number of classes.
In this setting, the variance term $\sigma_{c}^{w}(x)^2$ allows the model to quantify aleatoric uncertainty. 

\subsection{Uncertainty \& Noisy Labels in Remote Sensing}
\noindent Gaussian Processes \cite{rasmussen_gaussian_2006}, and Deep Gaussian Processes \cite{damianou_deep_2013} have been applied in several EO tasks \cite{camps-valls_survey_2016, camps-valls_perspective_2019}, such as biophysical parameter retrieval \cite{mateo-sanchis_gap_2018, svendsen_deep_2020} and RS image classification \cite{morales-alvarez_remote_2018}.
While these models offer quality estimates of uncertainty, their complexity poses challenges in training, making their utilization in big Earth data problems computationally infeasible.
Lately, more scalable BNNs have been proposed for uncertainty quantification in EO. 
In particular, Monte Carlo (MC) Dropout \cite{gal_dropout_2016} has been employed for biophysical parameter retrieval \cite{martinez-ferrer_quantifying_2022}, ice and water detection \cite{asadi_evaluation_2020}, urban landscape object segmentation \cite{kampffmeyer_semantic_2016}, and aerial image segmentation \cite{dechesne_bayesian_2021}.
Moreover, Bayesian Deep Learning \cite{he_bayesian_2023} and VI-based Bayesian Convolutional Neural Networks (CNNs)  \cite{joshaghani_bayesian_2023} have been used for hyperspectral image classification.

Addressing label noise in RS has been the focus of several studies.
Heteroscedastic Gaussian Processes have been used for biophysical parameter retrieval in cases where noise is signal-dependent \cite{lazaro-gredilla_retrieval_2014}.
Additionally, task-specific methodologies have been explored.
Li et al. \cite{li_learning_2019} proposed a multi-view CNN framework that iteratively corrects label errors, while Jiang et al. \cite{jiang_hyperspectral_2019} introduced a random label propagation algorithm to cleanse label noise for hyperspectral image classification. 
To enhance model resilience, a categorical cross-entropy loss function has been employed to train CNNs robustly for RS image classification \cite{li_improved_2022}.
Xu et al. \cite{xu_dual-channel_2022} proposed a dual-channel residual network combined with a noise-robust loss function to reduce the influence of mislabeled samples.
Noisy-tolerant methods have also been used for high-impact applications like drought detection \cite{jordi_droughts}, where a label correction method relying on model outputs was used.
Other strategies include complementary learning and deep metric learning with a robust softmax loss to mitigate the adverse effects of noisy labels \cite{li_complementary_2022}.
Methods for addressing multi-label noise in RS include collaborative learning approaches \cite{aksoy_multi-label_2022}, generative reasoning strategies \cite{sumbul_generative_2023}, and pseudo-labeling \cite{mirpulatov_pseudo}.

\section{Methodology}
\label{sec:method}

\noindent Existing approaches for modeling label noise in EO are often task-specific and lack general applicability. 
Furthermore, the relationship between label noise and the inherent aleatoric uncertainty remains largely unexplored in the field. 
In this work, we follow the work of Collier et al. \cite{collier_simple_2020} that proposed the adoption of a flexible probabilistic framework capable of modeling heteroscedastic label noise and estimating heteroscedastic aleatoric uncertainty.
The method builds upon the standard heteroscedastic classification model of Kendall and Gal \cite{kendall_what_2017} (See Eq.~\ref{eq:tau=1}).
In this section, we provide a brief overview of the method for completeness and refer readers to the original work for more details.

\subsection{Probabilistic ML framework}
\label{sub:prob_ml}
\noindent The method considers a latent variable generative process for the labels.
The generative process is handled by a latent variable $u_{c}(x)$, which is associated with each class $c$ and input $x$.
This variable is the sum of a deterministic vector $f^w_{c}(x)$ and an unobserved stochastic component $\epsilon_{c}$, expressed as: $u_{c}(x) = f^w_{c}(x) + \epsilon_{c}$.
A label is generated by sampling from $u_{c}(x)$ and taking the $\arg\max$ of all classes, i.e. class $c^*$ is the generated label if $u_{c}(x) \leq u_{c*}(x), \forall c \in 1,\ldots K$, where $K$ is the total number of classes. 
The probability $p_c (x)$ that an input $x$ belongs to class $c$, can be then expressed as
\begin{equation}
\begin{aligned}
  p_c(x) & = P(\arg\max_k u_{k}(x)=c) \\ 
         & = \int \mathbf{1} \left\{ \arg\max_k u_k(x) = c \right\} p(\epsilon_c) \, d\epsilon_c
\end{aligned}
\label{eq:generative}
\end{equation}

Assuming that $\epsilon_{c}$ are independent and identically distributed (homoscedastic), after calculations, the probability $p_c(x)$ becomes the well-known softmax function used in standard ML classification:
\begin{equation*}
\begin{aligned}
    p_c(x) = \frac{\exp{u_c(x)}}{\sum_{k=1}^K\exp(u_{k}(x))}
\end{aligned}
\end{equation*}

Thus, this generative process is implicitly assumed in the training of standard ML classifiers. 
However, under the presence of input-dependent label noise, the assumption of identically distributed $\epsilon_{c}$ becomes restrictive, as the noise source varies from sample to sample and across different classes.
In such cases, it is necessary to account for differing levels of stochasticity for each sample (heteroscedasticity).
For instance, in land-use classification, images containing a mix of agricultural and arable land often present higher labeling uncertainty than more clear classes, such as forests.

To handle this heteroscedasticity, the method introduces a dependency between the noise terms $\epsilon_{c}$, the input, and classes, breaking the identically distributed assumption. 
Specifically, $\epsilon_{c} \sim \mathcal{N}(0, \sigma^w_{c}(x)^2)$, where $\mathcal{N}$ denotes a Normal distribution, and $\sigma^w_{c}(x)^2$ models noise levels that vary based on the input and class.
Under these assumptions, computing $p_c(x)$ in Eq. ~\ref{eq:generative} is intractable.
However, it can be approximated using a temperature-scaled softmax and MC sampling.
Moreover, to enable gradient-based optimization, $u_{c}(x)$ can be reparameterized as $u_c(x) = f_c^w(x) + \sigma _c^w(x)\mu_c$, where $f_c^w(x)$ and $\sigma _c^w(x)$ are deterministic components and $\mu_c\sim\mathcal{N}(0,1)$.
The final calculation of $p_c(x)$ is obtained as:
\begin{equation}
\begin{aligned}
    p_c(x) & = P(\arg\max_k u_{k}(x)=c) \\
           & \approx \mathbb{E}_{\epsilon_{k}\sim\mathcal{N}(0,\sigma^{w}_{c}(x)^2)}\left[\frac{\exp{(u_{c}(x)/\tau)}}{\sum_{k=1}^K\exp {(u_{k}(x)/\tau)}}\right], \tau > 0 \\
           & \approx \frac{1}{S}\sum_{s=1}^{S}\frac{\exp((f_c^{w}(x) + \sigma_{c}^{w}(x)\mu_{c}^{s})/\tau)}{\sum_{k=1}^{K}\exp((f_{k}^{w}(x) + \sigma_{k}^{w}(x)\mu_{k}^{s})/\tau)},
\end{aligned}
\label{eq:tempered-softmax}
\end{equation} 
where S is the number of MC samples.

In this setting, the variance term $\sigma_{c}^{w}(x)^2$ is critical allowing the model to: i) discern label noise varying across individual samples, learning to assign higher values to inputs exhibiting higher noise and lower values to those with less noise; and ii) provide aleatoric uncertainty for the predictions of each sample.
The practical implementation of this solution is simple.
The network's final layer is trained to predict both $f_c^w(x)$ and $\sigma_c^w(x)$, representing the mean and variance of a Normal distribution for each class.
The final predictions are then obtained using the formula in Eq.~\ref{eq:tempered-softmax} and the uncertainties as discussed in Sec. \ref{sub:uncertainty}.
This operation is computationally efficient because the sampling is confined to the final layer of the network, unlike, for instance, BNNs, which involve sampling across all network parameters. 
The temperature parameter $\tau$ plays a critical role in balancing the trade-off between the approximation bias and variance in the MC gradient estimates.
Its optimal value requires separate computation for each task using a validation set. 
Notably, when $\tau = 1$, the method converges to the standard heteroscedastic classification model (Eq.~\ref{eq:tau=1}).

\subsection{Aleatoric Uncertainty Estimation}
\label{sub:uncertainty}

\noindent As previously discussed, the term $\sigma_{c}^{w}(x)^2$ enables the estimation of aleatoric uncertainty for the predictions.
In this study, to quantify this uncertainty, we use the MC samples from the predictive distribution in Eq. \ref{eq:tempered-softmax} and compute their variance. 
Specifically, for a given input $x$, an individual MC sample is expressed as: $$p_{c,s} = \frac{\exp((f_c^{w}(x) + \sigma_{c}^{w}(x)\mu_{c}^{s})/\tau)}{\sum_{k=1}^{K}\exp((f_{k}^{w}(x) + \sigma_{k}^{w}(x)\mu_{k}^{s})/\tau)}$$
and the mean prediction across MC samples as: $$\bar{p_c} = \frac{1}{S}\sum_{s=1}^{S}p_{c,s}$$
The aleatoric uncertainty for each class is computed as the variance of all drawn MC samples:
$$uncertainty_c = \frac{1}{S} \sum_{i=1}^S (p_{c, i} - \bar{p_c})^2.$$

\begin{figure*}[!t]
    \centering
    \includegraphics[width=1.0\linewidth]{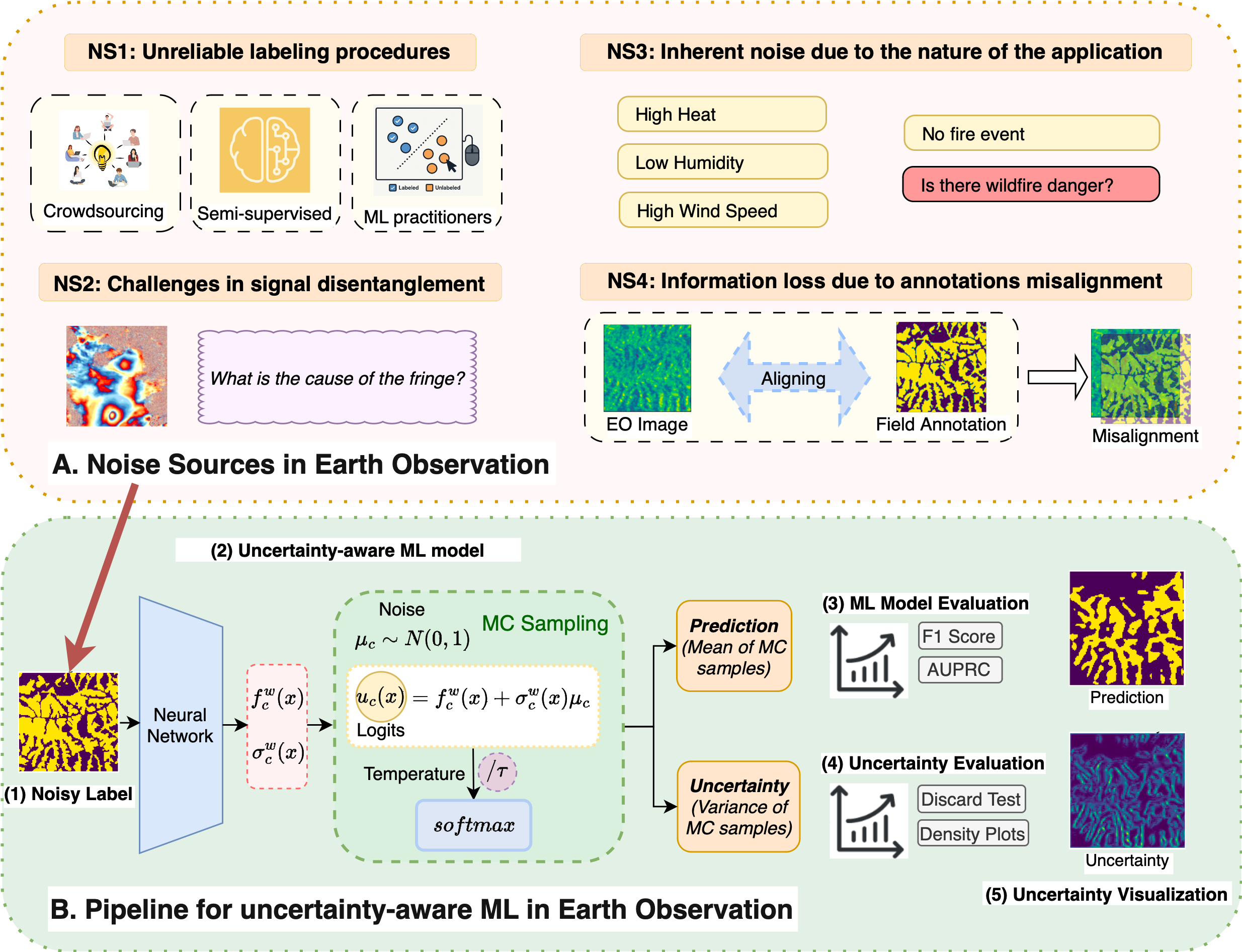}
    \caption{(A) The categorization of Label Noise Sources (NS) in Earth Observation (EO). (B) The uncertainty-aware Machine Learning (ML) pipeline for modeling label noise in EO that is proposed in this study. An uncertainty-aware ML model is used to model the label noise in EO. A Normal distribution is induced in the logits of a Neural Network, and the model is trained to predict its mean $f_{c}^{w}(x)$ as the output and $\sigma_{c}^{w}(x)$ as its heteroscedastic uncertainty. Monte Carlo (MC) sampling is used to generate multiple samples from this distribution, enabling the estimation of the final model prediction (mean of the samples) and its associated uncertainty (variance of the samples). A temperature parameter $\tau$ is used, scaling the logits for a tempered softmax calculation. Model performance is assessed using standard evaluation methods ($F_1$ score and Area Under Precision-Recall Curve (AUPRC)), while uncertainty estimates are assessed using dedicated uncertainty evaluation methods (Discard Test and Uncertainty Density plots) and visualizations.}
    \label{fig:method}
\end{figure*}

\section{Application to Earth Observation}

\noindent This section presents the application of the probabilistic framework in EO.
Sec. \ref{sub:prob_eo} and Fig. \ref{fig:method}-B summarize our pipeline for modeling label noise and estimating uncertainty in EO applications.
In Sec. \ref{sub:source} and Fig. \ref{fig:method}-A, we identify the unique sources of label noise commonly found in EO datasets, while Sec. \ref{sub:datasets} provides an overview of the diverse datasets and downstream applications used in this study to assess the framework's effectiveness.
Finally, Sec. \ref{sub:experiments} details the experimental setups for all tasks and the methods used to evaluate both predictive performance and uncertainties reliability.

\subsection{Investigation of Probabilistic ML in Earth Observation}
\label{sub:prob_eo}

\noindent Building upon the methodology outlined in Sec. \ref{sec:method}, we investigate DL modeling under label noise in EO tasks.
We employ standard deterministic DL models and integrate the probabilistic module at the logits level (Fig.~\ref{fig:method}-B(2)).
MC sampling is used to generate both the model predictions and the associated uncertainty estimates.
Leveraging the framework's flexibility, we develop uncertainty-aware DL models for a range of high-stakes EO applications, exhibiting variations in input modalities, ML configurations, and noise sources.
These applications span single-label and multi-label image classification, as well as image segmentation tasks. 
To assess the effectiveness of the probabilistic model, we benchmark its performance against standard deterministic DL models and the standard heteroscedastic model.
Moreover, we introduce a dedicated pipeline for uncertainty-aware ML modeling in EO.
This pipeline emphasizes the assessment of uncertainty estimates in addition to traditional performance evaluation, highlighting that the trustworthiness of a model is often more important than merely improving its performance.

The key steps of our pipeline are summarized in Fig. \ref{fig:method}: 
(1) We define the primary label noise sources in EO supervised datasets and use datasets with noisy labels representing each category; (2) we train uncertainty-aware DL models using the probabilistic framework of Sec. \ref{sec:method}; (3) we evaluate model performance, validating that uncertainty-aware models maintain or improve predictive accuracy compared to standard DL models; (4) we assess the reliability of uncertainty estimates to ensure their trustworthiness in EO tasks, and (5) we use the estimated uncertainties to generate visualizations of sample data, demonstrating how these uncertainties can be interpreted to support decision-making in real-world applications.




\begin{table*}[htbp]
    \centering
    \caption{Overview of the applications, datasets, noise sources, input modalities and Machine Learning (ML) tasks used in this study. SAR refers to Synthetic Aperture Radar data and InSAR to Interferometric Synthetic Aperture Radar data.}
    \begin{tabular}{ccccc}
         Application & Dataset & Noise Source & Input Modality & ML Setup \\
         \midrule
         Land Use Land Cover Scene Classification & BigEarthNet \cite{sumbul_bigearthnet_2019} & 1 & Optical Images & Multi-class multi-label image classification \\
         \midrule
         Landslide Segmentation & Landslides Dataset \cite{boehm_deep_2022} & 4 & SAR & Image segmentation\\
         \midrule
         Volcanic Activity Detection & Hephaestus \cite{bountos_hephaestus_2022} & 2 & InSAR & Image classification \\
         \midrule
         Wildfire Danger Forecasting & Wildfires Dataset \cite{kondylatos_wildfire_2022} & 3 & Multi-modal & Time-series classification \\
    \end{tabular}
    \label{tab:my_label}
\end{table*}

\begin{figure*}[!t]
    \centering
    \includegraphics[width=1.0\linewidth]{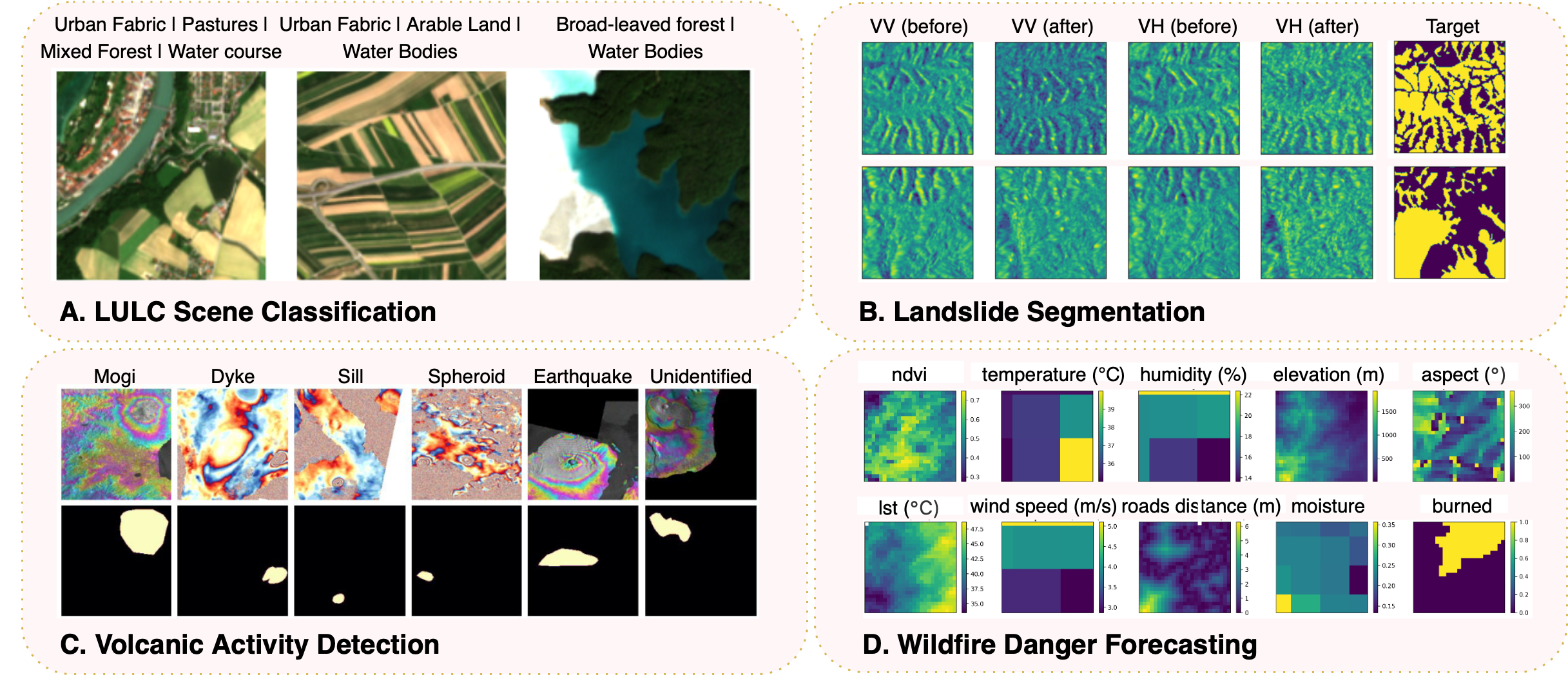}
    \caption{Examples of data samples from the datasets used in this study, highlighting sources of label noise. (A) BigEarthNet: The discrepancies in labeling strategies of the Corine Land Cover database introduce label noise. (B) Landslides dataset: Misalignments between in-situ annotated masks and earth observation images contribute to labeling inconsistencies. (C) Hephaestus: Atmospheric contributions and coherence variations challenge the annotation, leading to labels with heteroscedastic noise. (D) Wildfires dataset: Label noise stems from the stochastic nature of wildfire occurrence, where similar environmental conditions do not always lead to the same target class.}
    \label{fig:dataset}
\end{figure*}

\subsection{Label Noise Sources in Earth Observation}
\label{sub:source}
\noindent Annotating EO images poses challenges, as different annotation strategies introduce distinct sources of label noise.
Approaches such as semi-supervised labeling, non-expert annotations, or in-field data annotation each have their limitations.
Additionally, the inherent complexity of EO applications and the intrinsic ambiguity in satellite data interpretation further impact label quality.
In this work, we systematically define and categorize the key Noise Sources affecting labels in EO datasets as follows:

\begin{customenum}
\item \textbf{NS1: Unreliable labeling procedures.} In EO applications, labels are often generated through methods such as annotation by ML practitioners, crowdsourcing initiatives, or unsupervised/semi-supervised learning techniques. While these approaches offer a cost-effective way of acquiring labeled datasets, they are generally considered less reliable regarding label quality \cite{snow_cheap_2008}.
\item \textbf{NS2: Challenges in signal disentanglement.} Even when domain experts are involved in the annotation process, the risk of inducing noise in the labels remains considerable.
This may occur due to limited information provided to the experts or the complexity of distinguishing signals in satellite imagery.
Furthermore, in case multiple experts label the same sample, inconsistencies in labeling results often emerge \cite{raykar_learning_2010}.
\item \textbf{NS3: Inherent noise due to the nature of the application.} In certain EO tasks, label noise is an inherent consequence of the problem being addressed. 
This is particularly common in applications related to natural hazards and disaster management, where similar variable conditions may or may not lead to the occurrence of specific physical phenomena.
In these cases, the occurrence or non-occurrence of an event is inherently stochastic \cite{prapas_deep_2021}, making it challenging to assign clear-cut positive or negative labels.
\item \textbf{NS4: Information loss due to annotations misalignment.} An important factor contributing to label noise involves the loss of information during the dissemination of data \cite{jiang_hyperspectral_2019}. 
This occurs when expert labeling is not directly applied to the data but relies on alternative sources, such as in-situ measurements or field annotations.
In such cases, even if the annotation process is precise, the post-alignment between input-label pairs can introduce inconsistencies.
\end{customenum}


\subsection{Datasets and Applications}
\label{sub:datasets}

\noindent To evaluate the effectiveness of the probabilistic model in capturing the various sources of noise in EO datasets, we carefully select four diverse datasets, each exhibiting one distinct source of label noise.
This section provides an overview of the selected datasets and tasks considered, discussing the specific noise sources associated with each.

\subsubsection{Land Use Land Cover Scene Classification}
\noindent BigEarthNet \cite{sumbul_bigearthnet_2019} is a multi-class, multi-label dataset tailored for Land Use Land Cover (LULC) scene classification, consisting of Sentinel-2 optical image patches spanning 10 European countries. 
Each image patch in the dataset is annotated with one or more LULC class labels derived from the Corine Land Cover (CLC) database \cite{buttner_corine_2014} (Fig.~\ref{fig:dataset}-A). 
Despite its utility, BigEarthNet is subject to NS1, which stems from the methodologies used to create the CLC database.
The bottom-up approach employed in CLC implementation involves the integration of national databases at the European level.
Thus, the variations in annotation methodologies across different countries pose challenges to annotation consistency.
Particularly, some of the countries employ semi-automatic or automatic techniques blending national in-situ data with satellite image processing, while others adhere to standard visual interpretation.
These challenges are verified by empirical evaluations comparing CLC against field-based LULC mappings, indicating that the dataset's reliability falls below 90\% \cite{buttner_corine_2014}. 
Furthermore, these evaluations reveal the heteroscedasticity in label noise, demonstrating that certain classes, such as sparse vegetation, exhibit lower reliability compared to more consistent classes like rivers and lakes.

\subsubsection{Landslide Segmentation}
\noindent The dataset used for landslide segmentation \cite{boehm_deep_2022} comprises SAR input data paired with expert-generated landslide masks for earthquake-triggered landslides in the Hokkaido region of Japan. 
Figure~\ref{fig:dataset}-B illustrates examples of such pairs.
The dual polarisation SAR intensity images (VV and VH) are used to predict the landslide masks.
In this dataset, the annotations for landslide masks were meticulously crafted by domain experts during fieldwork. 
However, precisely overlapping the SAR images with in-situ measurements is challenging, leading to misalignment between the annotated masked landslides and their actual spatial positions within the feature space. 
Thus, pixels that should correspond to landslides may not, and vice versa, particularly along the boundaries of the landslides.
These misalignments compromise the accuracy of input-label pairs, making this dataset susceptible to NS4.
Notably, the degree of inconsistency varies between pixels and samples, meaning that label noise is highly dependent on the specific image region and landslide being considered.



\subsubsection{Volcanic Activity Detection}
\noindent The Hephaestus dataset \cite{bountos_hephaestus_2022} serves as a comprehensive global InSAR repository tailored for volcanic activity monitoring. 
It contains a diverse set of labels, including ground deformation caused by volcanic activity, a mask depicting the location of the detected deformation, the deformation type (e.g. Sill, Dyke, Mogi) and the presence of atmospheric contributions in the InSAR imagery. 
Some representative samples, the ground deformation type, and the respective ground truth mask can be seen in Fig.~\ref{fig:dataset}-C. 
A team of domain experts methodically executed annotation efforts for the Hephaestus dataset. 
They examined potential fringe patterns and, when necessary, used external sources such as Digital Elevation Models (DEMs). 
Nevertheless, despite the involved expertise, labeling this dataset poses challenges, primarily due to NS2.
Atmospheric contributions create patterns that mimic fringes caused by actual ground deformation, complicating accurate annotation.
Moreover, regions with high incoherence present significant challenges in detecting ground deformations, particularly during the early stages of volcanic unrest.
The noise variability across samples induces heteroscedasticity, making some samples inherently more challenging to annotate due to variations in fringe patterns and incoherence levels. 

\subsubsection{Wildfire Danger Forecasting}
\noindent The dataset used for the wildfire danger forecasting task \cite{kondylatos_wildfire_2022} is derived from a spatio-temporal datacube with a daily temporal resolution and $1km \times 1km$ spatial resolution, centered around Greece. 
The dataset integrates variables from multiple sources, including satellite-derived vegetation status, meteorological observations, ground geomorphology data, and indicators of human activity (Fig.~\ref{fig:dataset}-D).
These variables are used to predict the likelihood that a given sample will be affected by a wildfire the subsequent day.
To formulate this as a supervised ML task, samples are extracted from the spatio-temporal dataset and categorized into two classes: high and low danger.
A key assumption in the sampling methodology of the referenced study is that wildfire danger increases when an actual fire incident occurs (positives) and remains low when no fire has occurred (negatives). 
However, wildfire occurrence is inherently stochastic, and the absence of a wildfire event does not necessarily indicate low wildfire danger \cite{prapas_deep_2021}.
To mitigate this stochasticity, negative samples are drawn from days without fire incidents across the entire geographical domain.
Despite these efforts, label noise from NS3 remains a significant concern.
This noise is particularly pronounced in the negative class, where samples with high wildfire risk may be incorrectly labeled as "low danger" simply because no fire occurred. 
Heteroscedasticity arises due to the varying noise labels across samples. 
For instance, negative samples collected during high-heat summer periods, when fire danger likely exists despite no recorded fire, exhibit higher noise levels compared to those collected during winter, when conditions are generally unfavorable for fire ignition.

\subsection{Experiments}
\label{sub:experiments}
\noindent This section presents the experimental setups for each task and describes the evaluation methods used to assess both predictive performance and uncertainty estimation.

\subsubsection{Experimental Setup}
\label{sub:exp_setup}

\noindent For all four tasks, we follow the experimental setups outlined in the respective referenced papers, using the best-performing model architectures and hyperparameters for the DL models.
For the Hephaestus dataset, where no established experimental protocols have been published, we design a dedicated experimental setup and evaluate multiple architectures to assess the performance of the different approaches.

For the \textit{LULC scene classification} task, the dataset consists of $120 \times 120$ patches, with all bands of Sentinel-2 used as input. 
The task is framed as multi-class, multi-label image classification, where each class is predicted independently using the binary cross-entropy loss.
We use the pre-defined train, validation, and test splits from BigEarthNet and employ the ResNet-$50$ \cite{he2016deep} model architecture.

For the \textit{landslide segmentation} task, the dataset consists of $128 \times 128$ patches derived from dual polarisation SAR intensity images captured before and after landslide events.
We use several pre-event SAR images and a single post-event image.
The pre-event time series are mean-aggregated, and the VV and VH bands from both pre-and post-event images are combined into four input channels serving as input to the segmentation model.
Differing from the original work, we introduce a three-way split of the data---training (70\%), validation (20\%), and test (10\%)---to enable the tuning of the temperature parameter of the probabilistic framework.
We utilize the U-Net++ \cite{zhou_unet_2018} architecture with a ResNet-$50$ encoder and train the model using cross-entropy loss. 

For the \textit{volcanic activity detection} task, we utilize single InSAR frames, resampled to a uniform resolution of $1024\times1024$.
The task is formulated as a binary classification problem, where the presence or absence of ground deformation is predicted using the cross-entropy loss.
A spatial split is used for dataset partitioning, selecting 12 unique locations to serve as the test set, which contains 547 samples with ground deformation and 6990 samples without deformation.
The test set distribution is designed to reflect real-world conditions, simulating an operational early warning system.
We employ the ResNet-$18$, ResNet-$50$, and DenseNet$121$ \cite{huang2017densely} model architectures in our experiments.

The \textit{wildfire danger forecasting} task is framed as a binary time series classification problem using the cross-entropy loss, with one class indicating fire danger and the other its absence.
Following \cite{kondylatos_wildfire_2022}, we use a 2:1 ratio of negative (low fire danger) to positive (high fire danger) samples.
Data from 2009 to 2018 are used for training, data from 2019 for validation, and data from 2020 for testing. 
The model implementation is based on a Long Short-Term Memory (LSTM) \cite{hochreiter} network, using a $10$-day time series of input variables. 

For all tasks, the optimal temperature $\tau$ for the probabilistic framework is selected based on validation set metrics, exploring values from $0.1$ to $1$ in steps of $0.1$, and from $1$ to $10$ in steps of $1$.
During both training and testing, we use $1000$ MC samples to estimate the model's final predictions and associated uncertainties.
The performance of the probabilistic framework is compared against models trained with the standard heteroscedastic approach ($\tau = 1$) and standard deterministic DL models.

\subsubsection{Evaluation Methods}
\label{sub:evaluation}
\noindent In all the experiments, $F_1$ score and the Area Under the Precision-Recall Curve (AUPRC) are used as performance evaluation metrics. 
Beyond these standard metrics, a key component of our study is to assess the reliability of the models' uncertainty estimates.
In line with the principle that predictions with low uncertainty should be accurate and inaccurate predictions should exhibit high uncertainty \cite{mukhoti_evaluating_2019}, we employ two methods to assess uncertainty reliability: the Discard Test \cite{haynes2023creating} and Uncertainty Density Plots \cite{stahl_evaluation_2020}. 

The \textit{Discard Test} evaluates the quality of uncertainty estimates by iteratively removing batches of the most uncertain predictions from the test set and measuring the model's error on the remaining samples. 
This process is repeated iteratively until all samples have been discarded, producing a curve that illustrates how the model error changes with each iteration.
A model that produces reliable uncertainty estimates should exhibit decreasing error as more uncertain predictions are discarded, indicating that more uncertain samples correspond to less accurate predictions.
In this study, we use the loss as an indicator of error and consider $10$ discard fractions.
The results of the Discard Test are visualized using a line plot that displays the discard fraction alongside the model's error, along with two key metrics: Monotonicity Fraction (MF) and Discard Improvement (DI).
MF measures how often model performance improves as more uncertain samples are discarded.
It is computed as:
$$MF = \frac{1}{N_{f}-1}\sum_{i=1}^{N_{f}-1}I(\epsilon_{i}\geq\epsilon_{i+1}),$$
where $I$ is the indicator function, $\epsilon_i$ is the model error (here the loss) at discard fraction $i$, and $N_f$ denotes the total number of considered discard fractions.
An MF value of $1$ indicates perfect monotonicity.
DI quantifies the average reduction in model error as the discard fraction increases and is given by:
$$DI=\frac{1}{N_{f}-1}\sum_{i=1}^{N_{f}-1}I(\epsilon_{i} - \epsilon_{i+1}).$$

The \textit{Uncertainty Density Plots} illustrate the distribution of uncertainty scores for test set samples, distinguishing between correctly and incorrectly classified instances.
The median uncertainty for each group is also reported.   
In a reliable model, misclassified samples are expected to exhibit higher uncertainty, whereas correctly classified samples should show lower uncertainty, leading to a clear separation between the distributions.

\begingroup

\renewcommand{\arraystretch}{2.0} 
\begin{table*}[ht!bp]
\centering
\caption{Comparison of model performance using the probabilistic Machine Learning (ML) model \cite{collier_simple_2020}, versus the standard heteroscedastic model \cite{kendall_what_2017} and deterministic Deep Learning (DL) models for Land Use Land Cover (LULC) scene classification, landslide segmentation, and wildfire danger forecasting. The optimal temperature scaling parameter $\tau^*$ is reported for each dataset. The highest-performing values for each dataset and metric are highlighted in bold.}
\resizebox{\textwidth}{!}{%
  \begin{tabular}{ccccccc}
    \toprule
    \multirow{2}{*}{Model Type} &
     \multicolumn{2}{c}{LULC scene classification ($\tau^* = 1$)} &
      \multicolumn{2}{c}{Landslide Segmentation ($\tau^* = 2$)} &
      \multicolumn{2}{c}{Wildfire Danger Forecasting ($\tau^* = 0.2$)} \\
      \cline{2-7}
     & {$F_1$ Score} & {AUPRC} & {$F_1$ Score} & {AUPRC} & {$F_1$ Score} & {AUPRC} \\
    \midrule
    Deterministic DL & 0.748 & 0.816 & 0.545 & \textbf{0.660} & 0.767 & 0.872 \\
    \hline
    Standard heteroscedastic model ($\tau = 1$) & \textbf{0.754} & \textbf{0.829} & 0.476 & 0.648 & \textbf{0.777} & 0.879 \\
    \hline
    Probabilistic ML ($\tau = \tau^*$) & \textbf{0.754} & \textbf{0.829} & \textbf{0.593} & 0.640 & 0.776 & \textbf{0.883} \\
    \bottomrule
  \end{tabular}%
  }
  \newline
  \label{tab:table_results}
\end{table*}
\endgroup

\begingroup

\renewcommand{\arraystretch}{2.0} 
\begin{table*}[ht!bp]
\centering
\caption{Comparison of model performance using the probabilistic Machine Learning (ML) model \cite{collier_simple_2020}, versus the standard heteroscedastic model \cite{kendall_what_2017} and deterministic Deep Learning (DL) models for volcanic activity detection across different model architectures. The optimal temperature scaling parameter $\tau^*$ is reported for each architecture. The highest-performing values for each architecture and metric are highlighted in bold.}
  \begin{tabular}{ccccccc}
  \toprule
    & \multicolumn{6}{c}{Volcanic Activity Detection} \\
    \cline{2-7}
    \multirow{1}{*}{Model Type} &
     \multicolumn{2}{c}{ResNet-18 ($\tau^* = 5$)} & 
      \multicolumn{2}{c}{ResNet-50 ($\tau^* = 0.9$)} &
      \multicolumn{2}{c}{DenseNet121 ($\tau^* = 1$)} \\
     & {$F_1$ Score} & {AUPRC} & {$F_1$ Score} & {AUPRC} & {$F_1$ Score} & {AUPRC} \\
    \midrule
    Deterministic DL & \textbf{0.402} &\textbf{0.357} &  0.436 & 0.437 & 0.339  & 0.359 \\
    \hline
    Standard heteroscedastic model ($\tau = 1$) & 0.315 & 0.261 & 0.428 &  0.435 & \textbf{0.495} & \textbf{0.468} \\
    \hline 
    Probabilistic ML ($\tau = \tau^*$) & 0.387 & 0.304 & \textbf{0.518} &  \textbf{0.518} & \textbf{0.495} & \textbf{0.468} \\
    \bottomrule
  \end{tabular}
  \newline
  \label{tab:table_results_hephaestus}
\end{table*}
\endgroup

\section{Results \& Discussion}
\noindent In this section, we present and analyze the results of our study, focusing on both predictive performance and uncertainty reliability.

\subsection{Evaluation Performance of the ML models}
\noindent The evaluation metrics for LULC scene classification, landslide segmentation, and wildfire danger forecasting are summarized in Tab.~\ref{tab:table_results}, while results for volcanic activity detection are presented in Tab.~\ref{tab:table_results_hephaestus}. 
For all applications, the $F_1$ score and the AUPRC are reported separately.
For the binary classification tasks $F_1$ score is reported for the positive class, while for the multi-class datasets, the micro $F_1$ score is used.

Despite the differences in input modalities, ML setups, and sources of label noise, the probabilistic framework consistently improves performance across most tasks. 
Notably, it achieves the highest scores in LULC scene classification and wildfire danger forecasting and outperforms deterministic models in volcanic activity detection when using ResNet-$50$ and DenseNet$121$. 
It fails to surpass deterministic models in AUPRC for landslide segmentation, despite achieving a significantly higher $F_1$ score.
Additionally, it underperforms in both metrics for volcanic activity detection when using ResNet-$18$.

Furthermore, except for the $F_1$ score in wildfire danger forecasting, the probabilistic model outperforms the standard heteroscedastic model.
In some cases, the optimal $\tau$ is found to be $1$,  indicating that the probabilistic and heteroscedastic models coincide.
In general, the best $\tau$ varies across datasets and model architectures, emphasizing the need for its careful tuning for improved performance.

The decrease in AUPRC for landslide segmentation can be attributed to the increased variability introduced by the probabilistic model---which relies on multiple MC samples per prediction.
While this approach improves robustness to label noise, it can also reduce confidence in predictions, particularly in ambiguous regions such as landslide boundaries, where label noise is concentrated. 
This can result in fewer true positives identified with high confidence, leading to a lower AUPRC in this specific application. 
Since AUPRC evaluates performance across multiple thresholds, it is more sensitive to these confidence variations. 
In contrast, the $F_1$ score, computed using a fixed threshold, remains unaffected.

The most significant performance improvement is observed in the task of volcanic activity detection. 
This enhancement can be attributed to the task's inherent challenges, particularly the high similarity between positive and negative atmospheric InSAR signals, which introduces substantial uncertainty during labeling. 
In this context, the probabilistic approach has effectively captured this labeling uncertainty, resulting in a marked improvement in the performance metrics of the larger models i.e. ResNet-$50$ and DenseNet$121$.


\subsection{Uncertainties Evaluation}

\begin{figure}[t!]
  \centering
  \begin{subfigure}[t!]{0.49\linewidth}
  \includegraphics[width=1.0\linewidth]{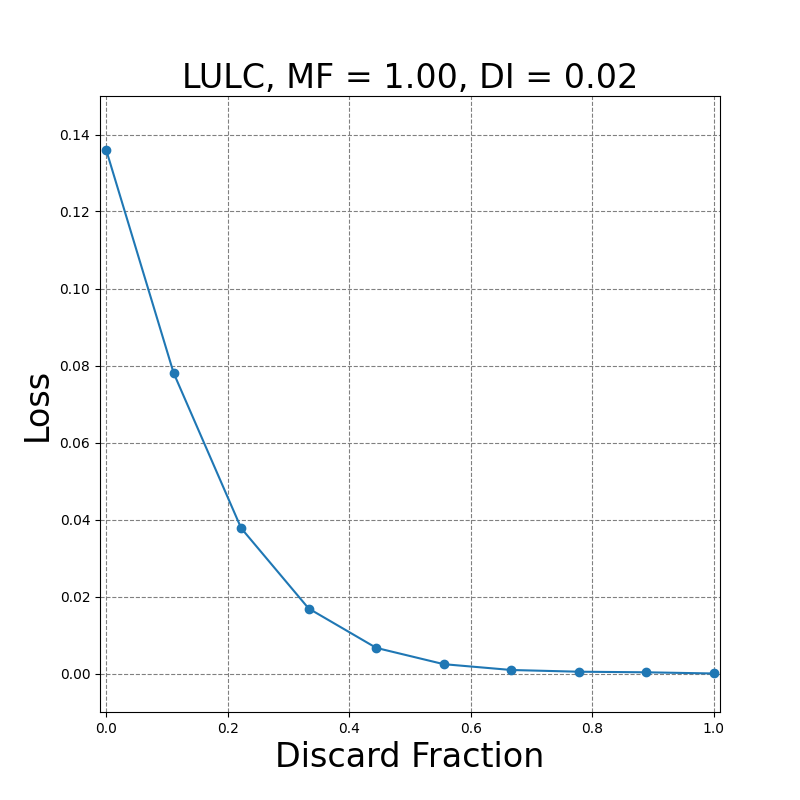}
  \end{subfigure}
  \hfill
  \begin{subfigure}[t!]{0.49\linewidth}
  \includegraphics[width=1.0\linewidth]{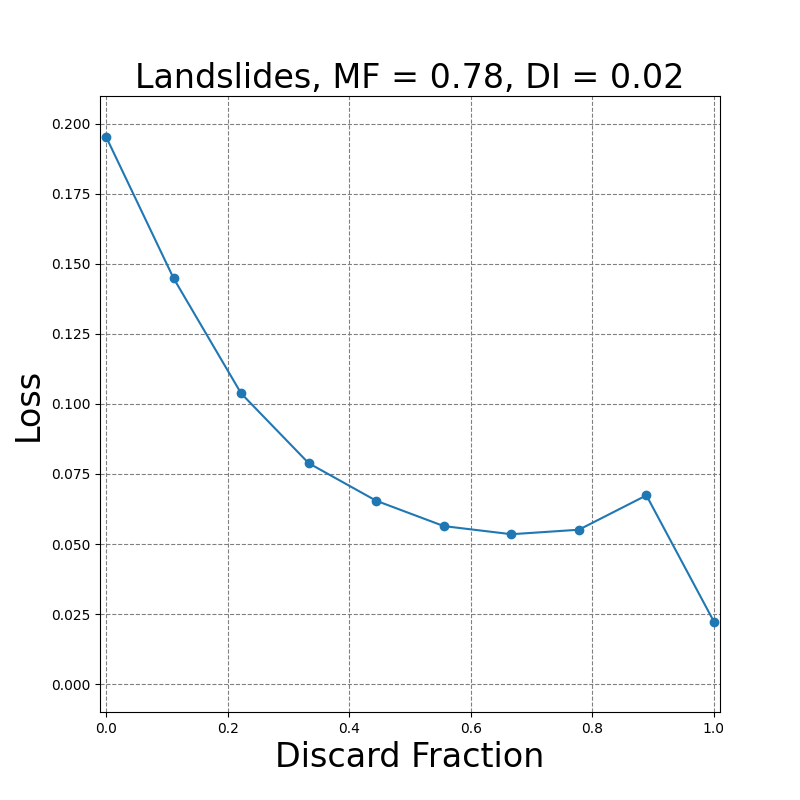}
  \end{subfigure}
  \begin{subfigure}[t!]{0.49\linewidth}
  \includegraphics[width=1.0\linewidth]{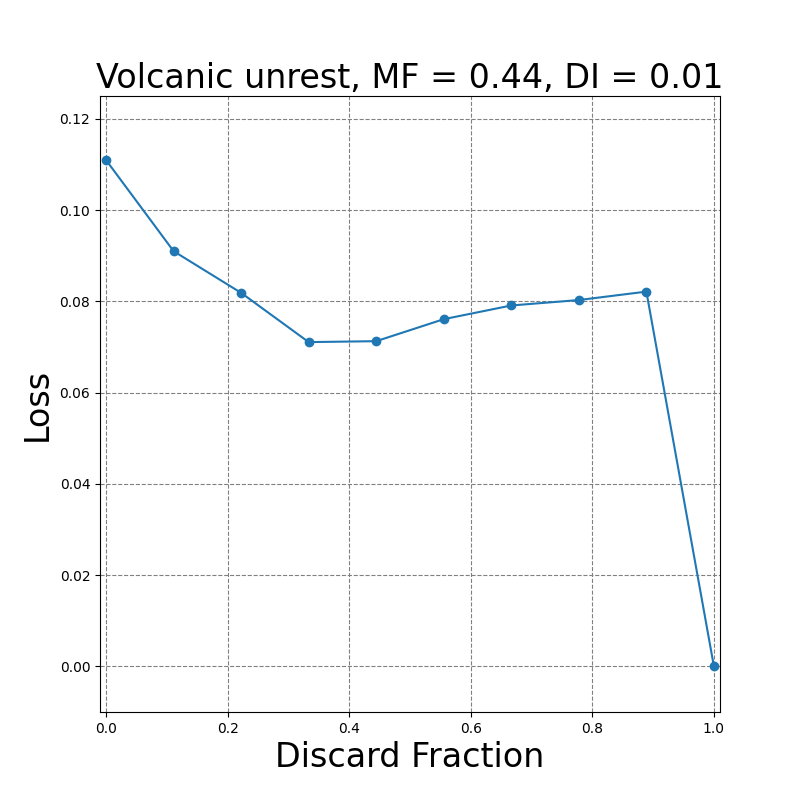}
  \end{subfigure}
  \begin{subfigure}[t!]{0.49\linewidth}
  \includegraphics[width=1.0\linewidth]{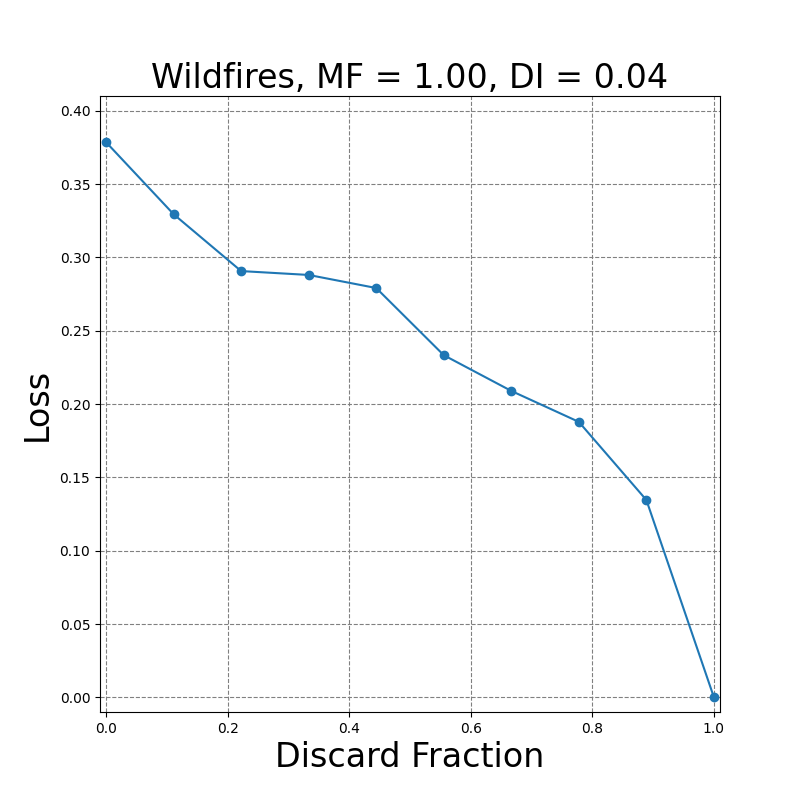}
  \end{subfigure}
  \caption{Discard test plots across all tasks. A reliable model should exhibit a decreasing error trend as the discard fraction increases, indicating that most uncertain samples correspond to higher loss values. The Monotonicity Fraction (MF) measures the frequency with which the error decreases upon discarding uncertain samples, while the Discard Improvement (DI) quantifies the average reduction in model error as the discard fraction increases. LULC refers to Land Use Land Cover.}
  \label{fig:discard-test}
\end{figure}

\begin{figure}[t!]
  \centering
  \begin{subfigure}[t!]{1.0\linewidth}
  \includegraphics[width=1.0\linewidth]{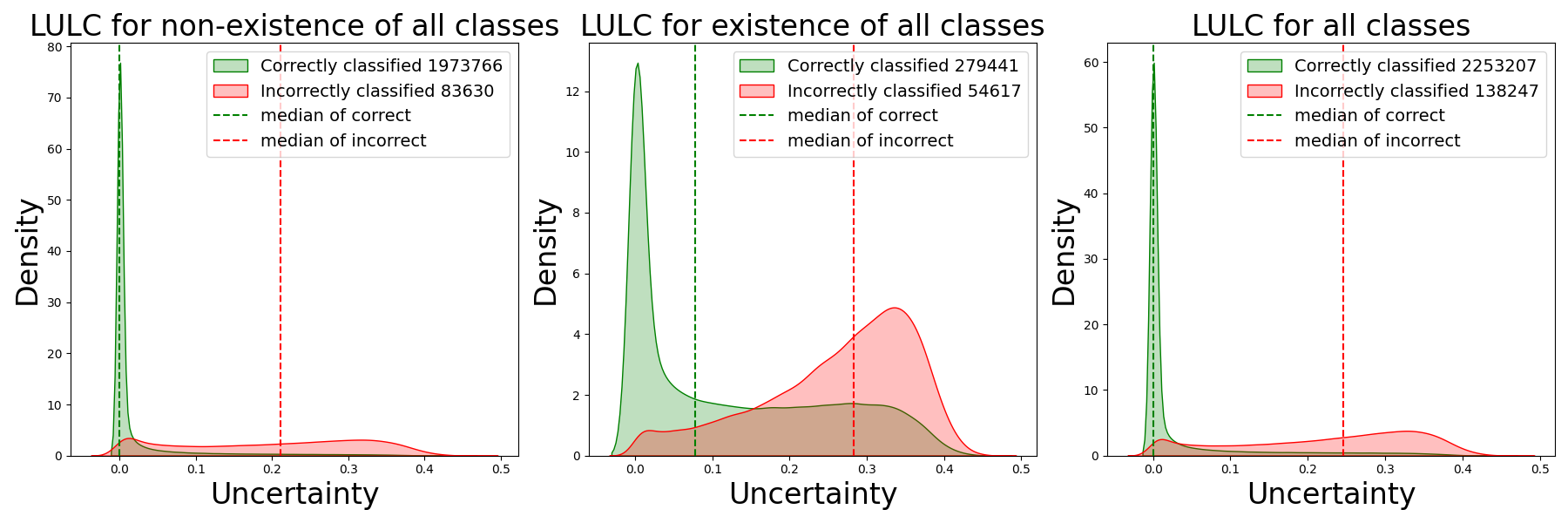}
  \end{subfigure}
  \hfill
  \begin{subfigure}[t!]{1.0\linewidth}
  \includegraphics[width=1.0\linewidth]{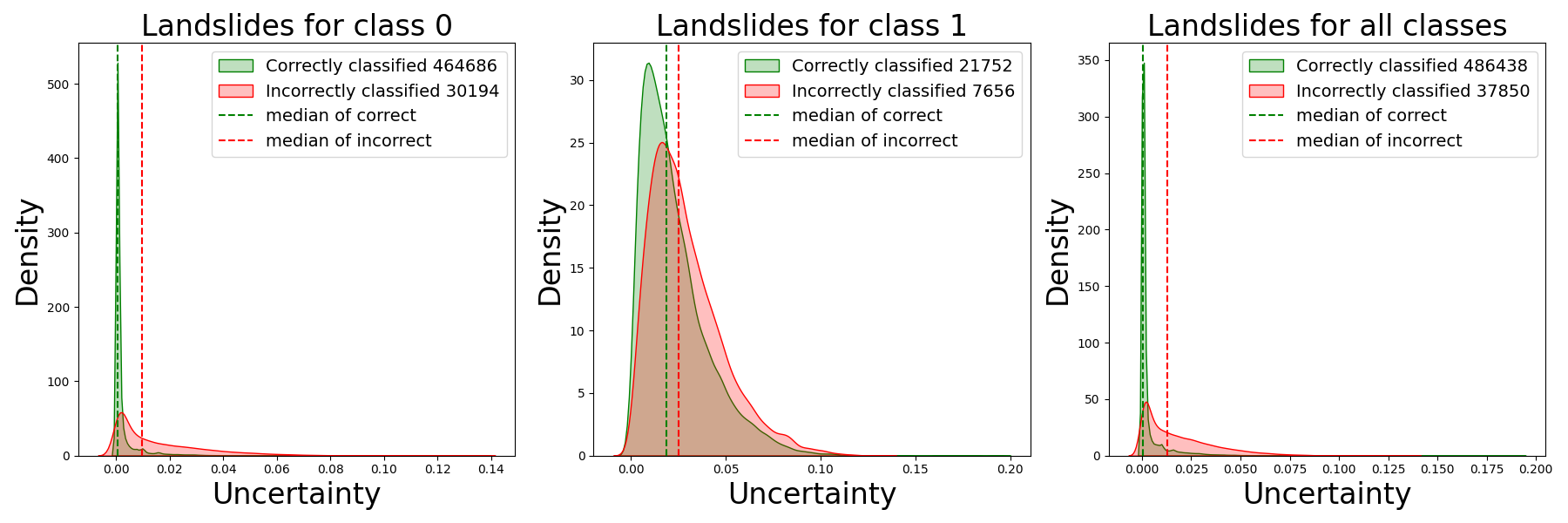}
  \end{subfigure}
  \begin{subfigure}[t!]{1.0\linewidth}
  \includegraphics[width=1.0\linewidth]{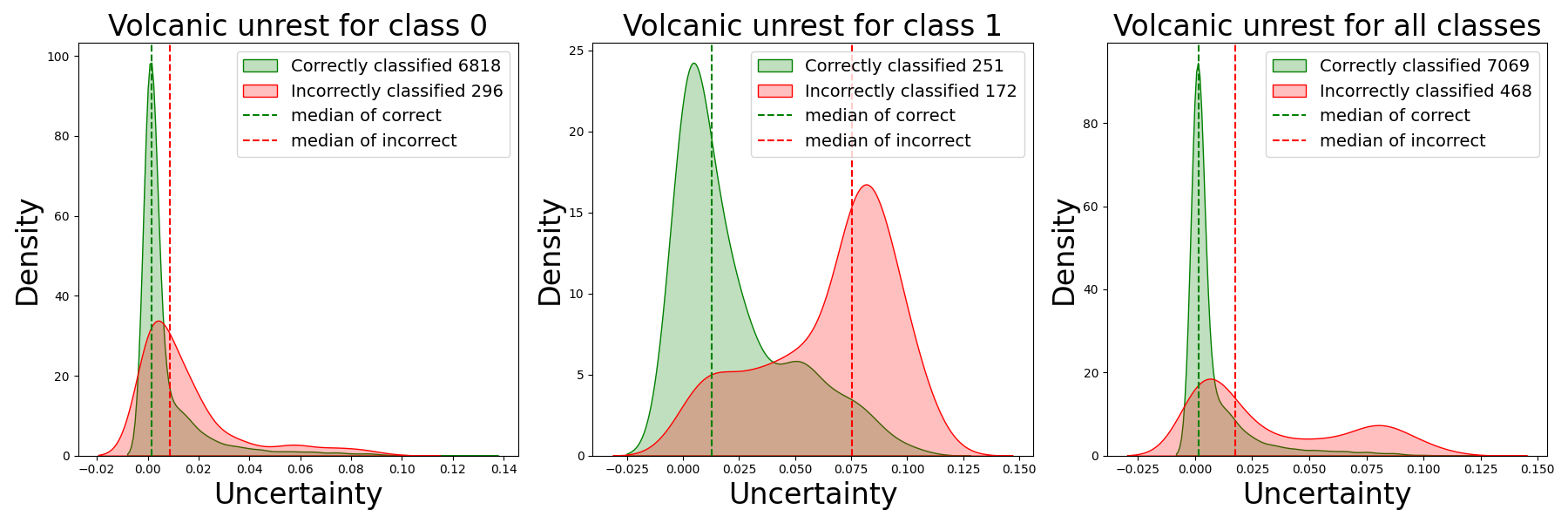}
  \end{subfigure}
  \begin{subfigure}[t!]{1.0\linewidth}
  \includegraphics[width=1.0\linewidth]{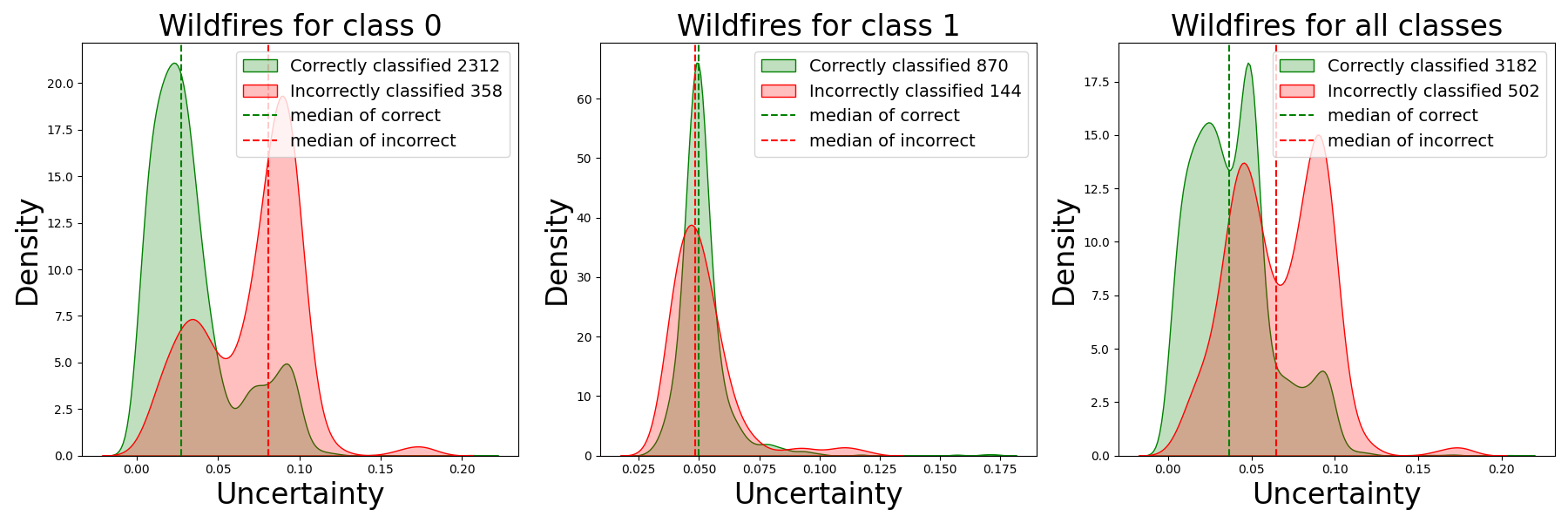}
  \end{subfigure}
  \caption{Uncertainty density plots across all tasks, presented for all classes combined and separately for the positive and negative classes. Vertical dashed lines indicate the median uncertainty for each group. Reliable uncertainty estimates are characterized by distinct, well-separated distributions with minimal overlap. LULC refers to Land Use Land Cover.}
  \label{fig:densities}
\end{figure}

\noindent Beyond optimizing predictive accuracy, this study primarily aims to evaluate the reliability of the uncertainty estimates produced by the probabilistic model. 
To achieve this, we assess the quality of these estimates using Discard Test plots (Fig.~\ref{fig:discard-test}) and Uncertainty Density plots (Fig.~\ref{fig:densities}).

\paragraph{Discard Test} Figure \ref{fig:discard-test} illustrates the Discard Test plots for all tasks, where the most uncertain samples are progressively removed, and the model's loss is calculated on the remaining samples.
Across all tasks, the discard test exhibits the optimal behavior, with model loss consistently decreasing as more uncertain samples are removed.
This indicates a clear alignment between high uncertainty and high-loss samples, highlighting the reliability of uncertainty estimates across the diverse applications in this study.
In volcanic activity detection, a slight increase in loss occurs after discarding more than half of the most uncertain samples (discard fraction $\geq 0.5$), before decreasing again in the last values.
This increase appears more like a plateau rather than a significant rise in loss. 
This behavior is likely due to the dataset's high imbalance, where most samples correspond to non-deformation cases that the model classifies with high confidence (low loss) and low uncertainty (See Fig. \ref{fig:densities}, third row).

Despite variations in loss functions and scales across tasks, DI remains positive, indicating a consistent overall reduction in loss.
Moreover, MF further supports these findings, reaching a perfect score ($1$) for LULC scene classification and wildfire danger forecasting while achieving $0.78$ for landslide segmentation. 
In volcanic activity detection, the observed increase in loss impacts MF, resulting in a lower score of $0.44$.

\paragraph{Uncertainty Density Plots} Figure \ref{fig:densities} presents the uncertainty density plots for all tasks, illustrating the distribution of the predicted uncertainties for correctly and incorrectly classified samples. 
For each application, three plots are provided.
In binary classification tasks, the first plot represents the uncertainty densities for negative labels, the second for positive labels, and the third for all. 
For the multi-label LULC task, the first plot shows the uncertainty densities for all labels assigned as $0$ (absent in the sample), the second for all labels assigned as $1$ (present in the sample), and the third combines them all.
This separation between negative and positive labels helps to reveal any class-specific variations in uncertainty reliability.

As shown in the "all classes" plots (third column), the probabilistic model consistently assigns higher uncertainty to misclassified instances, in line with the principle that predictions with lower uncertainty should be accurate, while inaccurate predictions should exhibit higher uncertainty. 
While this trend holds for all tasks, a more detailed analysis of the plots is conducted to extract task-specific insights.

For the LULC task, we conduct a deeper investigation of the impact of noise by providing separate uncertainty density plots for each class in the Appendix.
The results demonstrate that the observed trend persists across all classes, indicating high reliability in this task.

\begin{figure*}[t!]
  \centering
  \begin{subfigure}[t!]{1.0\linewidth}
    \includegraphics[width=1.0\linewidth]{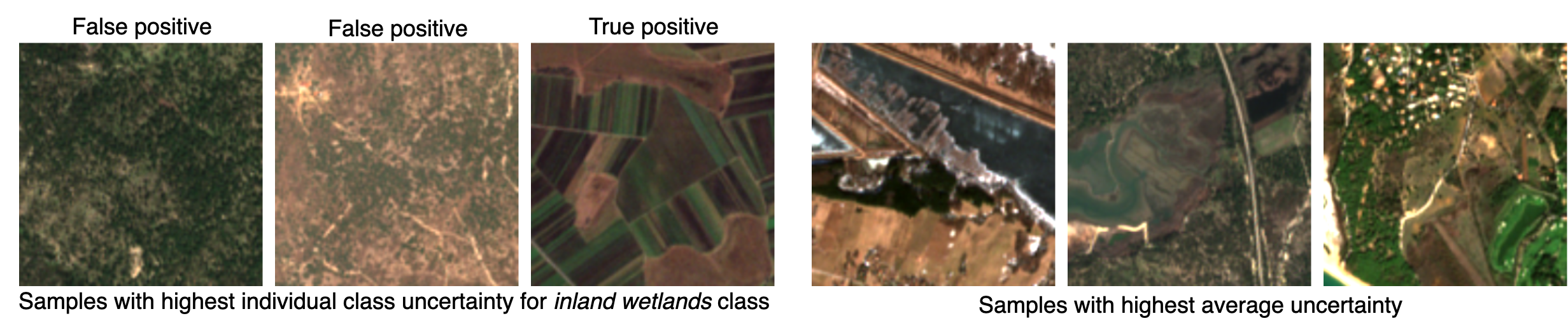}
    \caption{Examples of uncertain samples in the BigEarthNet dataset. The left panel shows samples with the highest uncertainty from the \textit{inland wetlands} class, which exhibits the greatest individual uncertainty among all classes. The right panel displays samples with the highest average uncertainty across all classes.}
    \label{fig:noise-a}
  \end{subfigure}
  \hfill
  \begin{subfigure}[t!]{1.0\linewidth}
    \includegraphics[width=1.0\linewidth]{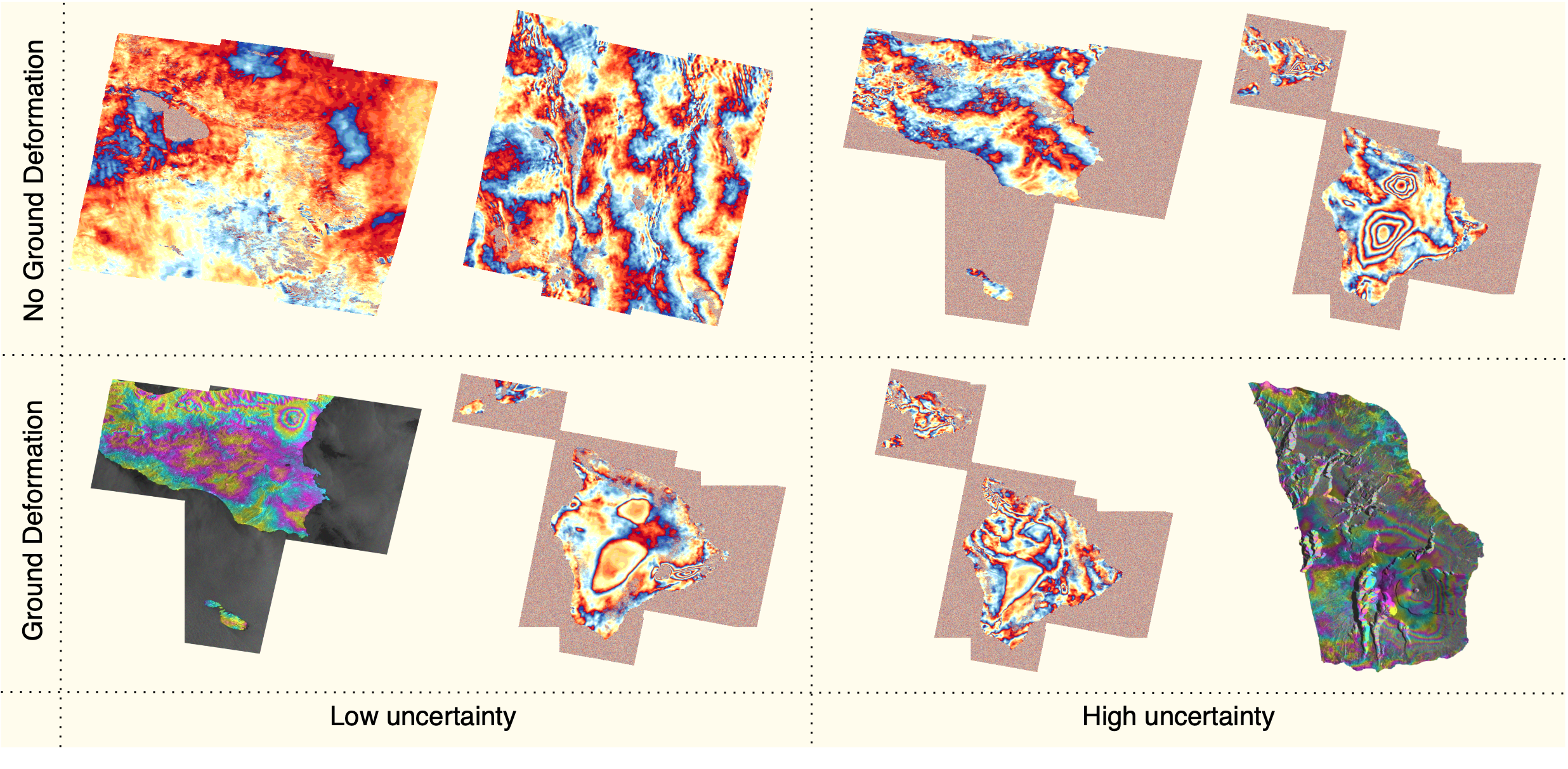}
    \caption{Samples from the Hephaestus dataset, categorized based on the presence of ground deformation and corresponding predicted uncertainty levels.}
    \label{fig:noise-b}
  \end{subfigure}
  \caption{Samples with low and high uncertainty for the volcanic activity detection and land use land cover scene classification tasks.}
  \label{fig:uncertainty}
  
\end{figure*}

\begin{figure*}[t!]
  \centering
  \begin{subfigure}[t!]{1.0\linewidth}
    \includegraphics[width=1.0\linewidth]{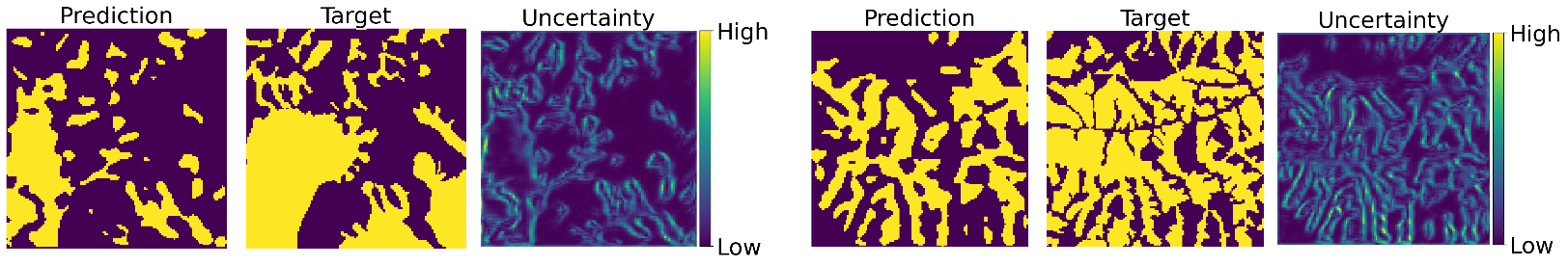}
    \caption{Ground truth and predicted segmentation maps for two landslide events, along with the predicted aleatoric uncertainty. The uncertainty is higher at the boundaries of the landslides, which represent the most challenging regions during manual annotation. }
    \label{fig:noise-c}
  \end{subfigure}
  \begin{subfigure}[t!]{1.0\linewidth}
    \includegraphics[width=1.0\linewidth]{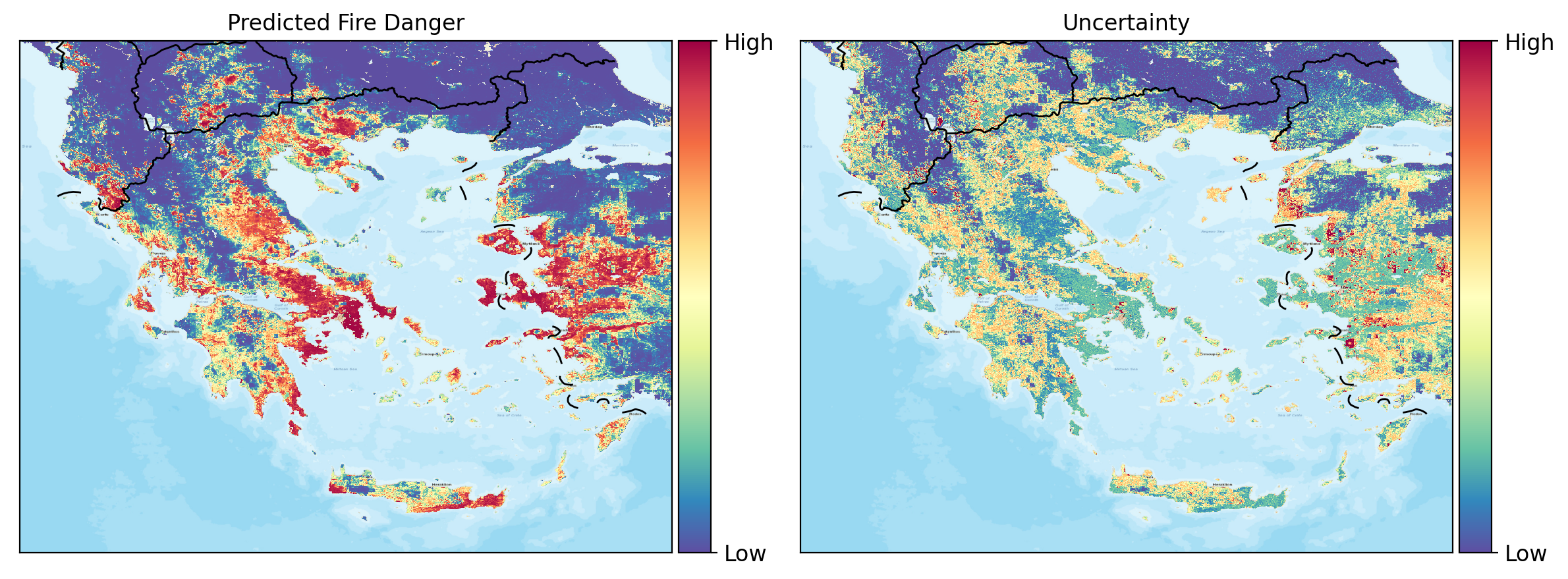}
    \caption{A wildfire danger prediction map accompanied by the predicted aleatoric uncertainty. These uncertainty maps can provide valuable insights for decision-makers, highlighting the confidence level of the machine learning models in their forecast.}
    \label{fig:noise-d}
  \end{subfigure}
  \caption{Prediction and uncertainty maps for the landslide segmentation and wildfire danger forecasting tasks.}
  \label{fig:uncertainty2}
\end{figure*}

In landslide segmentation, correctly classified samples from class $0$ (non-landslide areas) exhibit near-zero uncertainty, suggesting that these image regions are minimally affected by label noise.
In contrast, misclassified samples show higher uncertainty, a trend also reflected in the median values. 
Class $1$ (landslide areas) displays wider uncertainty distributions for both correctly and incorrectly classified samples.
This indicates that this class is more prone to noise and is more challenging for the model, as also illustrated in Fig. \ref{fig:noise-c}.
Despite these challenges, misclassified samples still exhibit higher uncertainty, highlighting the model's ability to quantify uncertainty reliably in this application.

In volcanic activity detection, a clear separation in uncertainty is observed for class $1$ (ground deformation).
For class $0$ (no ground deformation), uncertainty estimates are generally lower for both correctly and incorrectly classified samples, suggesting that this class exhibits less noise.
However, as indicated in the figure legend, only a small proportion of class $0$ samples is misclassified, implying that this class is relatively easy for the model to classify.
Despite smaller differences in uncertainty distributions for class $0$, the mode and median values for misclassified samples remain higher, showing the model's ability to distinguish uncertainty even in this class.


In wildfire danger forecasting, the uncertainty distributions are more complex. 
For class $0$ (low danger), while the separation between correct and incorrect classifications is clear, a bimodal distribution emerges for both groups.
One mode exhibits near-zero uncertainty, while the other lies in higher uncertainty values. 
This pattern reflects the seasonal composition of the dataset, where negative samples from different periods (e.g. summer vs winter) exhibit varying noise levels, which the model appears to have learned effectively.  
For class $1$ (high danger), overall uncertainty remains low, with the model struggling to distinguish between correct and incorrect classifications. 
This behavior aligns with the expectation that positive samples (indicating actual fire danger) exhibit lower noise levels. 
However, the persistence of low uncertainty even in the misclassified samples, suggests that the inherent characteristics of the data have limited the model's ability to express high uncertainty in this application, even for incorrect predictions.



\subsection{Qualitative Assessment of Aleatoric Uncertainty}

\noindent Figures ~\ref{fig:uncertainty} and ~\ref{fig:uncertainty2} present representative samples from all datasets, demonstrating how the predicted uncertainty estimates can provide additional insights for each task.

Figure~\ref{fig:noise-a} presents examples from the BigEarthNet dataset with the highest predicted uncertainty.
The three images on the right correspond to the samples with the highest average uncertainty across all classes. 
These samples exhibit multiple visually ambiguous classes, showing that the model is less confident when providing predictions for samples with complex or overlapping land cover types.
The three images on the left depict samples where the model predicted with high uncertainty the presence of inland wetlands, the class with the highest individual uncertainty among all.
This class is absent from the ground-truth labels in the first two images but present in the third.
This difficulty can be attributed to the inherent challenges in distinguishing inland waters in EO images, where spectral overlap with vegetation, forest, or bare soil leads to ambiguous classification, complicating even visual identification \cite{inland_PALMER20151}.

Figure~\ref{fig:noise-c} presents the predicted aleatoric uncertainty alongside segmentation maps for two landslide scenarios. 
The model consistently assigns higher uncertainty at the edges of the landslides, where annotation errors are more likely to occur.

Figure~\ref{fig:noise-b} presents eight samples from the Hephaestus dataset, categorized by ground deformation presence and predicted uncertainty levels.
In low-uncertainty samples, ground deformation is easily distinguishable, with InSAR data exhibiting distinct fringes, while no deformation regions appear smooth with minor perturbations, likely due to atmospheric effects. 
High-uncertainty samples, however, are more challenging. 
In the first no deformation sample, the source of uncertainty is unclear, whereas in the second, a prominent fringe likely confused the model, leading to lower-confidence predictions.
For ground deformation cases, in the first sample, the deformation pattern is subtle, justifying the high-uncertainty prediction. 
In the second, the InSAR is poorly overlaid on a DEM, distorting the output and confusing the model.
Interestingly, since Hephaestus maps InSAR data to pixel space, visualization choices, such as colormap selection and DEM overlays, can influence predictions by altering the physical meaning of the data.

Figure~\ref{fig:noise-d} presents a wildfire prediction map alongside its predicted aleatoric uncertainty.
Uncertainty is high in regions with intermediate wildfire danger probabilities, reflecting the model’s ambiguity in these cases, while it remains low for extreme probabilities near $0$ and $1$.
The observed relationship between uncertainty and fire danger probabilities suggests that softmax probabilities could serve as a reasonable proxy for data uncertainty in this task. 
However, further investigation is required to validate this hypothesis, which lies beyond the scope of the current study.


\section{Role of uncertainty in decision-making}

\noindent Trustworthiness is a fundamental requirement for the successful deployment of DL models in operational environments.
In this study, we leveraged uncertainty-aware models to address this critical need, demonstrating their reliability across various disaster management scenarios. 
Incorporating uncertainty can improve decision-making by providing valuable insights into the confidence levels of model predictions. 
For instance, in volcanic activity monitoring, uncertainty predictions can improve eruption preparedness by guiding decisions related to evacuations or ongoing surveillance, depending on the confidence level of the model.
Similarly, in wildfire danger forecasting, uncertainty-informed predictions allow for more strategic resource allocation by directing firefighting efforts to areas with higher confidence, which can optimize operations and reduce costs.
For landslide segmentation, uncertainty estimation can guide emergency responders in prioritizing search and rescue operations by focusing on areas with lower prediction uncertainty. 
Overall, by integrating uncertainty estimation, DL EO models transition from "black-box" systems to more transparent and trustworthy tools, ultimately supporting more robust and reliable decision-making in real-world applications.


\section{Conclusion}
\noindent In this work, we demonstrated the potential of uncertainty-aware ML in addressing input-dependent label noise in EO.
Through four high-stakes EO applications, particularly in disaster management, we highlighted how these approaches not only improve model performance but, more importantly, enhance the reliability of DL models in EO, an essential requirement for deploying ML solutions in real-world scenarios.
Our proposed pipeline, ranging from model performance assessment to the evaluation of uncertainty estimates and their interpretation in decision-making processes, holds significant potential for broader applications across various EO domains that require uncertainty quantification. 
A key limitation of this work is its exclusive focus on aleatoric uncertainty estimation. 
Future work will extend the proposed framework to include both aleatoric and epistemic uncertainty, further advancing the development of more robust, interpretable, and reliable ML-driven solutions in EO.

\section{Acknowledgments}
\noindent This work has received funding from the European Union's Horizon 2020 Research and Innovation Projects DeepCube and TREEADS, under Grant Agreement Numbers 101004188 and 101036926 respectively.

\clearpage
\bibliographystyle{ieeetr}
\bibliography{bibliography}

\appendix

\begin{figure*}[!t]
    \centering
    \includegraphics[width=0.99\textwidth]{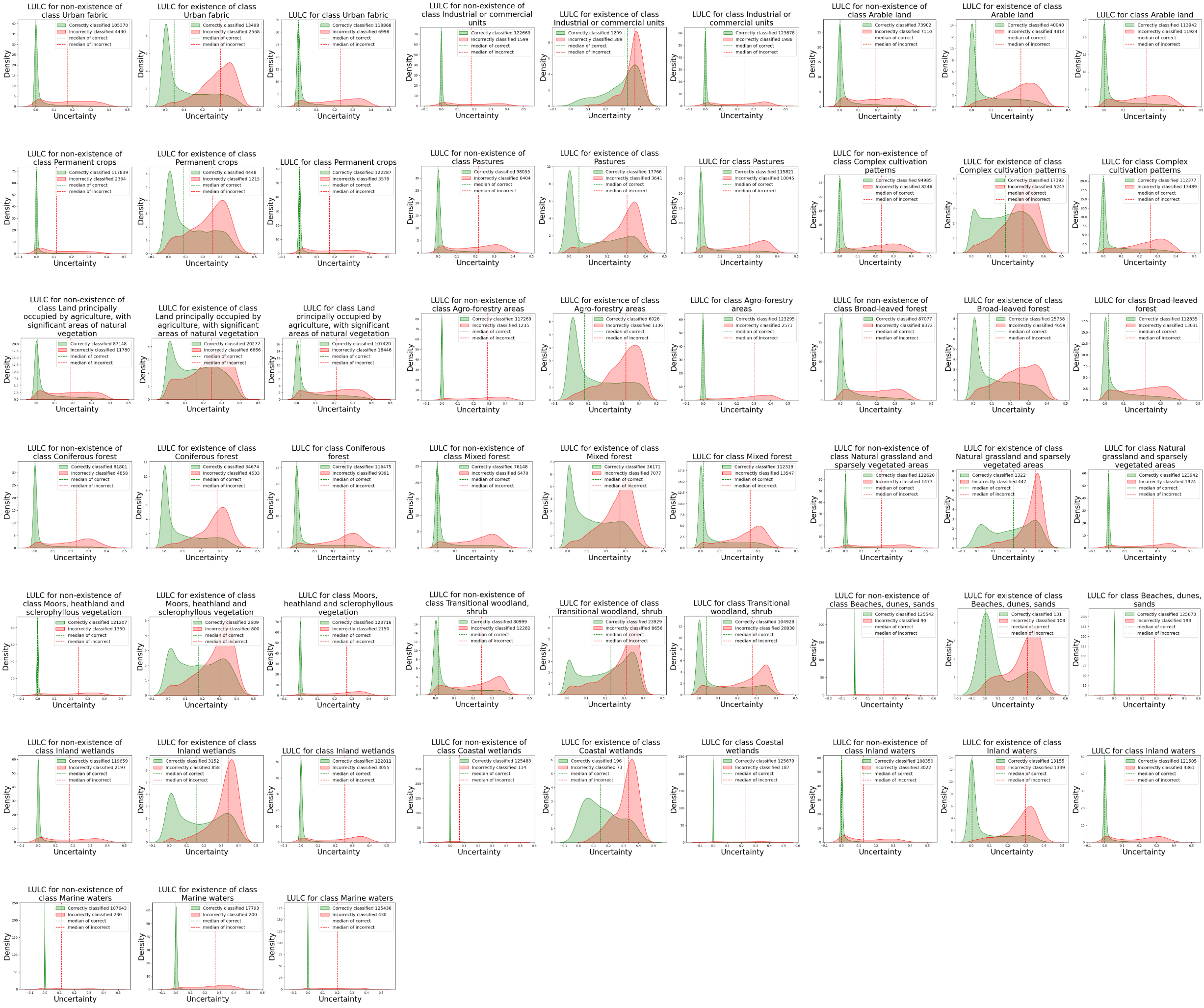}
    \caption{Uncertainty density plots for each class in the BigEarthNet dataset, applied to the Land Use Land Cover scene classification task. The first plot represents the samples where the class is absent, the second samples where the class is present, and the third includes all samples. Vertical dashed lines show the median uncertainty for each group. Reliable uncertainty estimates are characterized by distinct, well-separated distributions with minimal overlap.}
\end{figure*}

\end{document}